\pdfoutput=1

\documentclass[11pt]{article}

\usepackage{emnlp2021}

\usepackage{times}
\usepackage{latexsym}

\usepackage{breqn}
\usepackage{amsmath}
\usepackage{url}
\usepackage{subcaption}
\usepackage{makecell}

\usepackage[T1]{fontenc}

\usepackage[utf8]{inputenc}

\usepackage{microtype}
\usepackage{comment}

\usepackage{soul}
\usepackage{epstopdf}
\usepackage{booktabs}
\usepackage{graphicx}

\title{Exploring the Value of Pre-trained Language Models \\ for Clinical Named Entity Recognition}

\author{
         Samuel Belkadi$^{ *}$,  Lifeng Han$^{ *}$, Yuping Wu, 
          \and Goran Nenadic\\
         Department of Computer Science, The University of Manchester, UK \\ 
         * \textit{co-first authors}
         \\ {\tt samuel.belkadi@student.manchester.ac.uk} 
         \\
         {\tt 
         lifeng.han, yuping.wu, g.nenadic@manchester.ac.uk}         }

\begin{document}
\maketitle{}

\begin{abstract}
The practice of fine-tuning Pre-trained Language Models (PLMs) from general or domain-specific data to a specific task with limited resources, has gained popularity within the field of natural language processing (NLP).
In this work, we re-visit this assumption and carry out an investigation in clinical NLP, specifically Named Entity Recognition on drugs and their related attributes. 
We compare Transformer models that are trained from scratch to fine-tuned BERT-based LLMs namely BERT, BioBERT, and ClinicalBERT. Furthermore, we examine the impact of an additional CRF layer on such models to encourage contextual learning.
We use n2c2-2018 shared task data for model development and evaluations. 
The experimental outcomes show that 1) CRF layers improved all language models; 
2) referring to BIO-strict span level evaluation using macro-average F1 score, although the fine-tuned LLMs achieved 0.83+ scores, the TransformerCRF model trained from scratch achieved 0.78+, demonstrating comparable performances with much lower cost - e.g. with 39.80\% less training parameters; 
3) referring to BIO-strict span-level evaluation using weighted-average F1 score, ClinicalBERT-CRF, BERT-CRF, and TransformerCRF exhibited lower score differences, with 97.59\%/97.44\%/96.84\% respectively.
4) applying efficient training by down-sampling for better data distribution further reduced the training cost and need for data, while maintaining similar scores - i.e. around 0.02 points lower compared to using the full dataset.


Our models will be hosted at \url{https://github.com/HECTA-UoM/TransformerCRF}.
\end{abstract}

\section{Introduction}


Fine-tuning Pre-trained Language Models (PLMs) has demonstrated state-of-the-art abilities in solving natural language processing tasks, including text mining \cite{zhang-etal-2021-smedbert_TM}, named entity recognition \cite{2017neuroner}, reading comprehension \cite{sun-etal-2020-investigating_PK4Reading}, machine translation \cite{google2017attention,devlin-etal-2019-bert}, and summarisation \cite{gokhan-etal-2021-extractive_Fin_Summa,yuping_EDU_extractiveSum}. 
Domain applications of PLMs have spanned a much wider variety including financial, legal, and biomedical texts, in addition to traditional news and social media domains. 
For instance, experimental work on 
BioBERT \cite{10.1093/bioinformatics/btz682_BioBERT} and BioMedBERT \cite{chakraborty-etal-2020-biomedbert}
showcased high evaluation scores by exploiting BERT's \cite{devlin-etal-2019-bert} structure to train on biomedical data. Additionally, fine-tuned SciFive, BioGPT, and BART models produced reasonable experimental outputs on biomedical abstract simplification tasks \cite{li2023largePLABA}.

However, ongoing investigations try to understand the extent to which fine-tuning PLMs increases performances against language models trained from scratch on domain-specific tasks \cite{10.1145/3458754_PubMedBERT}. 
Researchers often assume that fine-tuning becomes indeed helpful when dealing with tasks that have limited available data, and where PLMs can leverage additional knowledge acquired from extensive out-of-domain or domain-related data.
Therefore, an important question arises: Given a domain-specific task, how limited should the available data be for mixed-domain pre-training to be considered beneficial? 
Surprisingly, no previous studies have provided statistics in this regard. In this paper, our focus is on clinical domain text mining, and our objective is to examine the aforementioned hypothesis. Specifically, we aim to determine whether PLMs outperform models trained from scratch when given access to limited data in a constrained setting, and to what extent this improvement occurs.


In comparison to other domains, clinical text mining (CTM) is still considered a relatively new task for PLM applications, as CTM is well known for data-scarce issues due to a small amount of human-annotated corpora and privacy concerns. 
In this work, we fine-tune PLMs from the \textit{general} domain BERT, \textit{biomedical } domain BioBERT, and \textit{clinical} domain ClinicalBERT, examining how well they perform on clinical information extraction task, namely drugs and drug-related attributes using n2c2-2018 shared task data via adaptation and fine-tuning. We then compare their results with ones of a lightweight Transformer model trained from scratch, and further investigate the impact of an additional CRF layer on the deployed models.


Section \ref{sec_RelatedW} gives more details on related works, Section \ref{sec_model_design} introduces the methodologies for our investigation, 
Section \ref{sec_experimental_setup} describes our data-preprocessing and experimental setups,
Section \ref{sec_experimental_eval} presents the evaluation results and ablation studies,
Section \ref{sec_discussion} further discusses data-constrained training looking back at n2c2-2018 shared tasks; 
finally, Section \ref{sec_conclude} concludes this paper and opens ideas for future works.
Readers can refer to the Appendix for more details on experimental analysis and relevant findings.


\section{Related Work}
\label{sec_RelatedW}

The integration of pre-trained language models into applications within the biomedical and clinical domains has emerged as a prominent trend in recent years. 
A significant contribution to this field is BioBERT \cite{10.1093/bioinformatics/btz682_BioBERT}, which was among the first to explore the advantages of training a BERT-based model from domain-specific knowledge, i.e. using biomedical data.
BioBERT demonstrated that training BERT using PubMed abstracts and PubMed Central (PMC) full-text articles resulted in superior performances on Named Entity Recognition (NER) and Relation Extraction (RE) tasks, within the biomedical domain.

However, since BioBERT was pre-trained on general-domain data such as Wikipedia or BooksCorpus \cite{7410368_booksCorpus} and then \textit{continuously-trained} on biomedical data, PubMedBERT \cite{10.1145/3458754_PubMedBERT} further examined the advantages of training a model from scratch solely on biomedical data, employing the same PubMed data as BioBERT to avoid influences of mixed domains. This choice was motivated by the observation that word distributions from different domains are represented differently in their respective vocabularies.
Furthermore, PubMedBERT created a new benchmark dataset named BLURB covering more tasks than BioBERT and including the terms: disease, drug, gene, organ, and cell.

PubMedBERT and BioBERT both focused on biomedical knowledge, leaving other closely related domains such as the clinical one for future exploration.
Subsequently, \cite{alsentzer-etal-2019-publicly_bioclinicalBERT} demonstrated that ClinicalBERT, trained using generic clinical text and discharge summaries, exhibited superior performances on medical language inference (i2b2-2010 and 2012) and de-identification tasks (i2b2-2006 and 2014). Similarly, \cite{huang2019clinicalbert} found that ClinicalBERT trained on clinical notes achieved improved predictive performance for hospital readmission after fine-tuning on this specific task.

In our work on the clinical domain, we use the n2c2-2018 shared task corpus which provides electric health records (EHR) as semi-structured letters (their heading specifying drug names, patient names, doses, relations, etc., and the body describing the diagnoses and treatment as free text).
We aim to examine how fine-tuned PLMs perform against domain-specific transformers trained from scratch, at biomedical and clinical text mining.

Regarding the usage of Transformer models for text mining, \cite{Wu2021transformer-ICT} implemented the Transformer structure with an adaptation layer for information and communication technology (ICT) entity extraction. \cite{al2021arabic_Transformer_ner} proposed to add a CRF layer on top of the BERT model to carry out Arabic NER on mixed domain data, such as news and magazines. \cite{Yan_etal2019Transformer4NER} demonstrated that the Encoder-only Transformer could improve previous results on traditional NER tasks in comparison to BiLSTMs. Other related works include \cite{zhang_n_wang2019TransformerCRF_spokenLU,Gan_etal2021Transformer_zhNER,zheng-etal-2021-TransformerCRF-power-meter,wang2022zhForestDisease_TransformerCRF} which applied Transformer and CRF for spoken language understanding, Chinese NER, power-meter NER, and forest disease text. 

\begin{figure*}[!t]
\begin{center}
\centering
\includegraphics*[width=0.85\textwidth]{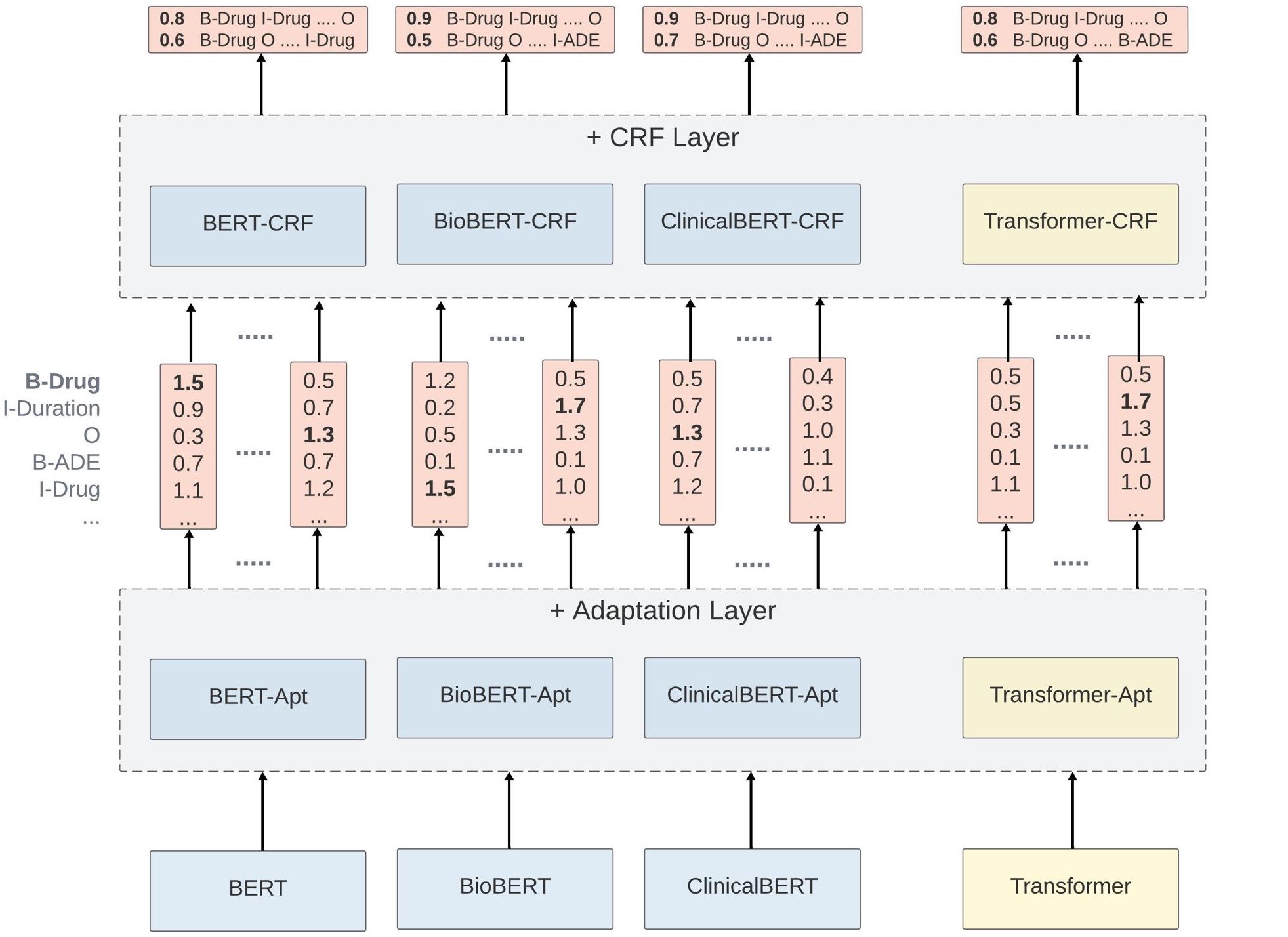}
\caption{Model Designs upon Investigations}
\label{fig:models-design-v2}
\end{center}
\end{figure*}

\section{Methodology and Experimental Designs}
\label{sec_model_design}
Figure \ref{fig:models-design-v2} displays the design of our investigation, which includes the pre-trained LLMs BERT \cite{devlin-etal-2019-bert}, BioBERT \cite{10.1093/bioinformatics/btz682_BioBERT}, and ClinicalBERT \cite{alsentzer-etal-2019-publicly_bioclinicalBERT}, in addition to an Encoder-only Transformer \cite{google2017attention} implementing the ``distilbert-base-cased'' structure and trained from scratch.

The first step is to adapt these models to Named Entity Recognition by adding an Adaptation (or Classification) layer, resulting in the following models: BERT-Apt, BioBERT-Apt, ClinicalBERT-Apt, and Transformer-Apt. This adaptation layer predicts probability distribution over all labels for each token independently.

Then, we compare the results of the above models with the same ones but implementing an additional Conditional Random Field (CRF) \cite{10.5555/645530.655813CRF2001} layer, obtaining BERT-CRF, BioBERT-CRF, ClinicalBERT-CRF, and Transformer-CRF models. Now, instead of independently predicting labels in a sequence, the CRF layer takes the neighbouring tokens with their corresponding labels to predict the label of the token under study.

\section{Data Pre-processing and Experimental Setups}
\label{sec_experimental_setup}

In this section, we introduce the n2c2-2018 corpus we utilise for model training and evaluations, as well as model optimisation strategies, efficient training, and evaluation metrics. 

\subsection{Corpus and Model Setting}
\label{sub_sec_corpus_model_setting}

Regarding the dataset, 
we utilise the standard n2c2-2018 shared task data from Track-2 \cite{henry20202018_n2c2_task2}: Adverse Drug Events and Medication Extraction in Electric Health Records (EHRs) \footnote{\url{https://portal.dbmi.hms.harvard.edu/projects/n2c2-2018-t2/}}. 
We note that The World Health Organisation (WHO) \footnote{\url{https://www.who.int/}} defines ADE as ``an injury resulting from medical intervention related to a drug'', while the Patient Safety Network (PSNet) defines it as ``harm experienced by a patient as a result of exposure to a medication'' \footnote{\url{https://psnet.ahrq.gov/primer/medication-errors-and-adverse-drug-events}}.
The aim of this task is to investigate whether ``NLP systems can automatically discover drug-to-adverse event relations in clinical narratives''. Three sub-tasks under this track include Concepts, Relations, and End-to-End. 
Among these, the first task is to identify drug names, dosage, duration, and other entities; the second task is to identify relations of drugs with adverse drug events (ADEs) and other entities given gold standard entities; finally, the third task is identical to the second one, but involves entities that have been predicted by systems. 
In total, this track provides 505 annotated files on discharge summaries from the Medical Information Mart for Intensive Care III (MIMIC-III) clinical care database \cite{johnson2016mimic}, for which annotation was carried out by four physician assistant students and three nurses. The presence of drugs and ADEs is reflected by entity tags and their corresponding attributes.
The 505 files were originally divided by the n2c2-2018 track-2 into 303 for model training and 202 for testing.
We subsequently split the original training data in a 9-1 fashion at random, i.e. 30 files for validation and the other 273 files for training or fine-tuning. Regarding sentence counts, we counted 41.497 sentences in the training set, 4.536 in the validation set, and 30.614 in the testing set.


 There are 18 (9x2) special labels plus an ``O'' label indicating that no label is attributed. The special labels cover 9 categories of events and medications namely ADE, Dosage, Drug, Duration, Form, Frequency, Reason, Route, and Strength. 
 We use the BIO labelling format from CoNLL2003 shared task \cite{tjong-kim-sang-de-meulder-2003-introduction_conll2003} with \textit{B} denoting ``beginning'', \textit{I} denoting ``inner token'' of events/medications, and \textit{O} for no label. 

\subsection{Model Optimisation and Training}

\begin{figure}[!b]
\begin{center}
\centering
\includegraphics*[width=0.49\textwidth]{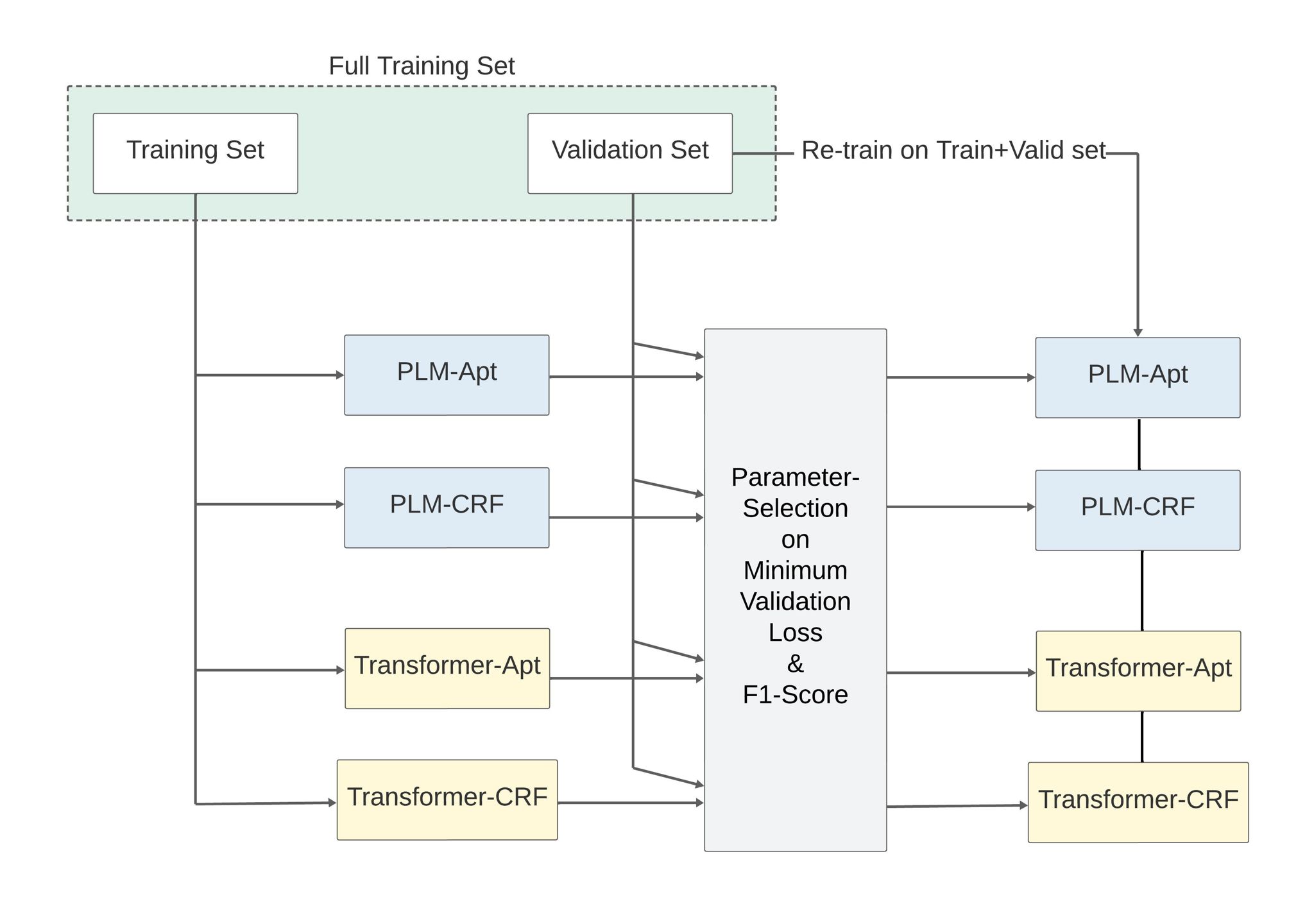}
\caption{Model Optimisation and Training}
\label{fig:model-selections-training-cropped}
\end{center}
\end{figure}

Figure \ref{fig:model-selections-training-cropped} displays the training procedure for each model. Firstly, we use the sub-training set for continuous learning on LLMs and for the training of TransformerApt and TransformerCRF. Then, the trained models are fed with our validation set for parameter optimisation based on F1 scores and minimum validation loss. 
Finally, the selected parameter sets for each model are used to re-train the models on the original n2c2-2018 training data consisting of both the sub-training and validation sets. As for foundation models, we use \textit{``bert-base-cased''}, \textit{``biobert-base-cased-v1.2''}, and \textit{``ClinicalBERT''} for LLMs; and \textit{``distilbert-base-cased''}'s architecture for the Transformer frame.

The number of trainable parameters of each model are listed in Table \ref{tab:parameters_comp}. It demonstrates that TransformerCRF reduces the amount of trainable parameters from BERT-CRF and BioBERT-CR by 39.80\%, and from ClinicalBERT-CRF by 49.51\%.

\begin{table}[!t]
\begin{center}
\centering
\begin{tabular}{ccc}
\toprule
 \multicolumn{1}{c}{CRF Model}     
                & Parameters   & \% to BERT-crf \\
\midrule
BERT-crf	& 108,325,282	& 1 \\
BioBERT-crf	& 108,325,282	& 1 \\
ClinicalBERT-crf	& 134,749,090 &	24.39\%$\uparrow$ \\
TransformerCRF	& 65,205,922	& 39.80\%$\downarrow$ \\ \hline \hline
 \multicolumn{1}{c}{Apt Model}     
                & Parameters   & \%to BERTApt \\
\midrule
BERT-Apt&	108,324,883	&1 \\
BioBERT-Apt	&108,324,883	&1 \\
ClinicalBERT-Apt	&134,748,691&	24.39\%$\uparrow$ \\
Transformer-Apt&	65,205,523&	39.80\%$\downarrow$  \\ 
\bottomrule
\end{tabular}
\caption{Number of trainable parameters of each model}
\label{tab:parameters_comp}
\end{center}
\end{table}

\begin{figure*}[!t]
\begin{center}
\centering
\includegraphics*[width=0.98\textwidth]{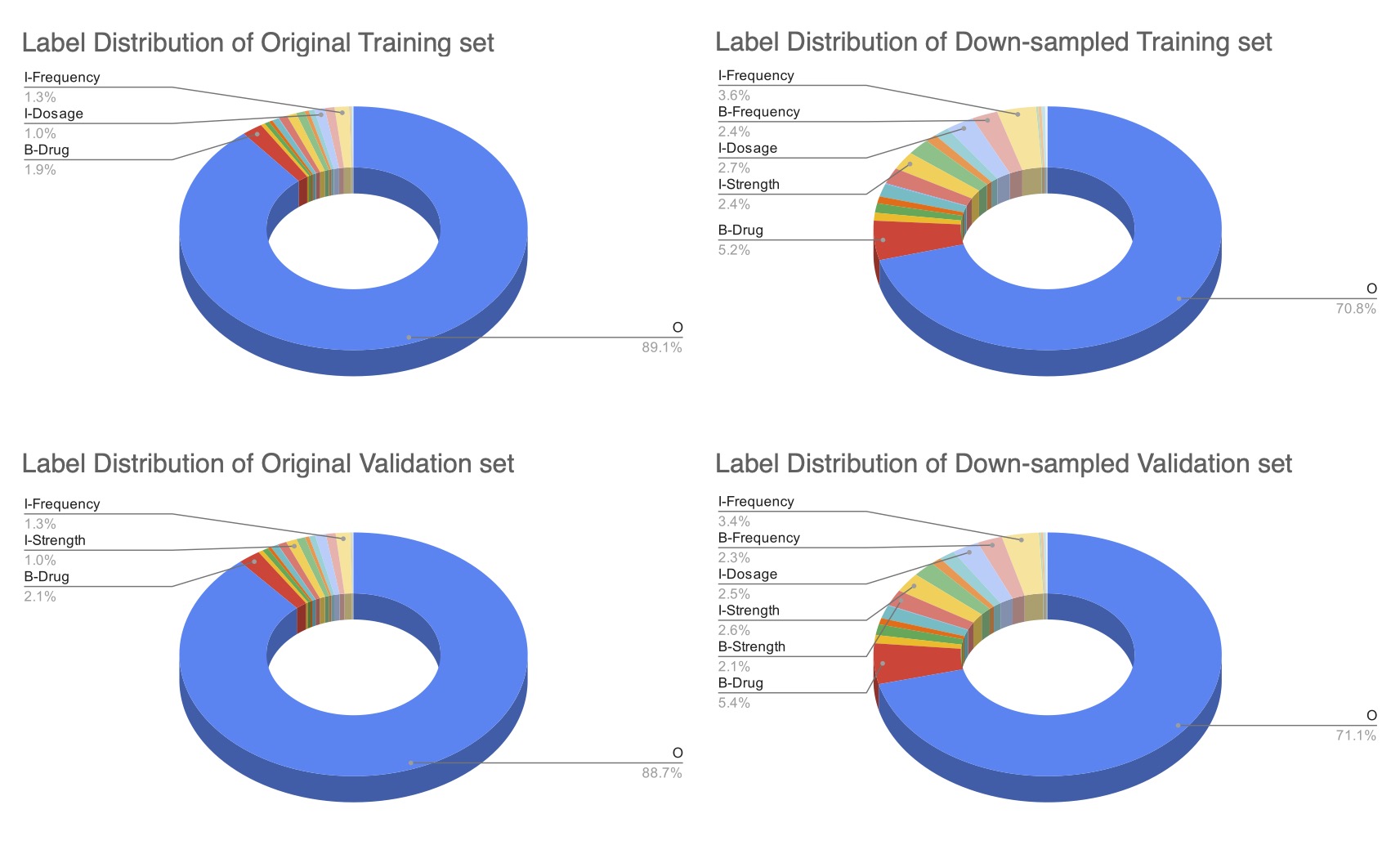}
\caption{Label Distributions in Training and Validation sets \textbf{before} (left) and \textbf{after} (right) down-sampling.}
\label{fig:n2c2_2018_label_dis_trainValid_w_SamBD}
\end{center}
\end{figure*}

\subsection{Efficient Training}
Because we observed that about 90\% of sentences in the original dataset were composed of only \textit{O} labels, we designed an efficient training step for this task by removing a certain amount of plain samples for better data distribution.
Specifically, we down-sample by removing empty-labelled sentences to approach a similar distribution with the CONLL2003 \cite{tjong-kim-sang-de-meulder-2003-introduction_conll2003} data, which has around 20\% of non-special label sentences. 
This optimisation reduces the proportion of (no-label) plain samples from approximately 75\% to 20\%, and the total number of tokens from about 41500 to 14000, which consequently lowers the computational cost of training.
A comparison of the label distributions between the original and optimised sets is displayed in Figure \ref{fig:n2c2_2018_label_dis_trainValid_w_SamBD}. Note that we leave the test set untouched.

In the original training set, most special labels have less than 1\% appearance, except for I-Frequency (1.3\%), I-Dosage (1.0\%), and B-Drug (1.9\%); However, in the down-sampled set, the special labels have increased frequencies - e.g. B-Drug (5.2\%) and I-Strength (2.4\%). Similar increases in special label frequencies are observed on the validation set when applying down-sampling.

Although employing the complete dataset may yield higher overall evaluation scores, our model might exhibit bias towards \textit{O} labels or be overly trained on an excessive number of such labels, thus not gaining any additional knowledge.
On the other hand, opting for the down-sampled set, which is significantly smaller and better distributed, allows for quicker hyperparameter selection, assuming that the achieved performance can be extrapolated to the original set. In instances where we observed slightly lower results, the disparity in scores was trivial in comparison to the resources saved by utilising a down-sampled dataset.

\subsection{Evaluation Metrics}

We calculate the macro- and micro-averaged precision, recall, and F1 scores, as well as the weighted scores. 
For multi-class labelling tasks, macro-averaging assigns the same weight to each label category, however, micro-averaging assigns the same weight to each sample token.
For weighted-average scores, the contribution of each class label is weighted by its size among all samples.

Because each event and medication can consist of multiple tokens (for example, \textit{"20 (B-strength) mg (I-strength) per (I-strength) day (I-strength)"}), we use BIO-level calculation to evaluate whether the model can distinguish the beginning and inner part of an event, and whether two distinct but consecutive events from the same class can be labelled separately as two different entities - i.e. the first token of the second event will be labelled \textit{B-xx} and not \textit{I-xx}. Additionally, we use token-level calculation to report whether entities are correctly identified without making any distinction between the beginning or inner part of an event.
It's worth noting that both span-level and token-level evaluations have been employed in the Multi-word Expression Prediction shared tasks of the ACL-SIGLEX group since 2017 \cite{Maldonado_han_moreau2017DetectionOV}.
\section{Experimental Outcomes}
\label{sec_experimental_eval}
In this section, we present both BIO-level (span-level) and Token-level (attribute-level) evaluation results on PLMs and Encoder-only Transformer, using the original full dataset.
Then we show the outcomes of using the Efficient Training method as an extension to our experimentation.


\subsection{BIO-Strict Evaluations}
The evaluation scores of each model trained on the full training dataset are depicted in Tables \ref{tab:macro-scores-test-set} and \ref{tab:weighted_n_micro_scores}. That includes macro-average, weighted average, and accuracy (micro-F1), for both the Classification-based and CRF-based models.

In terms of macro-average evaluations, ClinicalBERT-Apt achieved the highest Precision score at 85.30\% (with BioBERT-CRF coming in second at 84.26\%) while BERT-CRF secured the highest Recall score at 83.43\% (BioBERT-CRF being close again with 83.30\%) and the highest F1 score at 83.72\% (nearly identical to the one of ClinicalBERT-CRF). Notably, ClinicalBERT-CRF and BERT-CRF exhibit very similar F1 scores.
On the other hand, TransformerApt and TransformerCRF both attain Precision scores above 80\%, specifically 81.99\% and 82.51\%. However, we observe that their Recall scores fall below 80\%. Nevertheless, their performance remains competitive to fine-tuned Large Language Models, as they achieve F1 scores of 78.42\% and 78.74\% only exploiting the knowledge from the 303 training letters.

\begin{table}[!b]
\begin{tabular}{lccc}
\toprule
\multicolumn{1}{c}{Model} & P     & R     & F1    \\ 
\midrule \midrule
BERT-Apt                    & 0.835 & 0.831 & 0.832 \\
BERT-Crf                    & 0.842 & \textbf{0.834} & \textbf{0.837} \\
\midrule
BioBERT-Apt                 & 0.045 & 0.053 & 0.049 \\
BioBERT-Crf                 & \textit{0.843} & \textit{0.833} & 0.836 \\
\midrule
ClinicalBERT-Apt            & \textbf{0.853} & 0.810 & 0.825 \\
ClinicalBERT-Crf            & 0.850 & 0.829 & \textbf{0.837} \\
\midrule
TransformerApt              & 0.820 & 0.760 & 0.784 \\
TransformerCRF              & 0.825 & 0.762 & 0.787 \\
\bottomrule
\end{tabular}
\caption{BIO-strict macro-evaluation results. Highest scores are in bold and second highest in italic.}
\label{tab:macro-scores-test-set}
\end{table}

\begin{figure*}[!t]
\begin{center}
\centering
\includegraphics*[width=\textwidth]{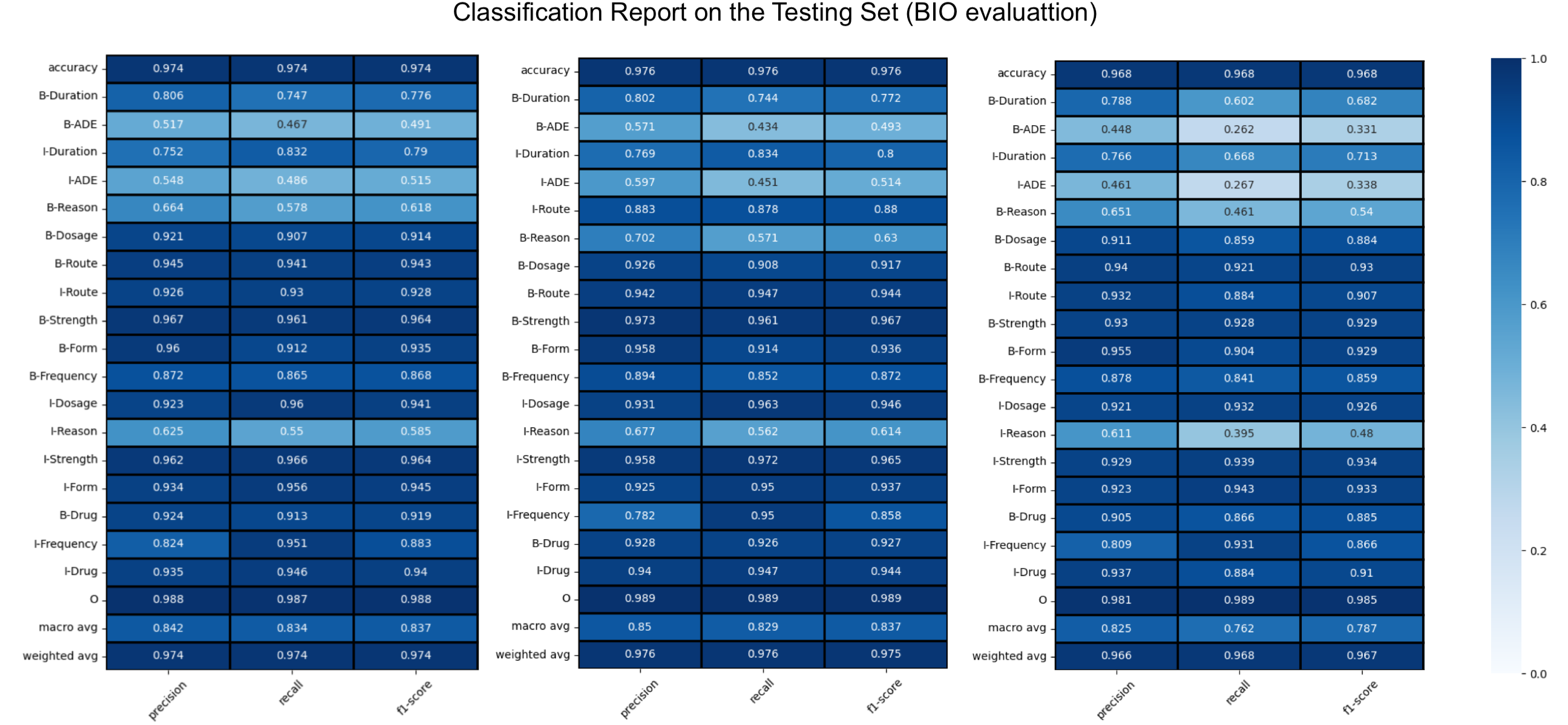}
\caption{BIO-strict Classification of BERT-CRF (left), ClinicalBERT-CRF (middle) and TrasformerCRF (right)}
\label{fig:CRF-BIO}
\end{center}
\end{figure*}

In the context of weighted and micro-average evaluations, as illustrated in Table \ref{tab:weighted_n_micro_scores}, ClinicalBERT-CRF emerges as the top performer across all columns, encompassing weighted precision, recall, F1, and accuracy. ClinicalBERT-Apt secures the second-highest positions in terms of weighted Recall and model accuracy, while BioBERT-CRF achieves the second-highest spots in weighted precision and F1 scores.
On the whole, when examining the weighted and micro evaluation scores, there isn't a substantial distinction between each model, with most of them falling within the score ranges of (96.60\% - 97.55\%) for weighted F1 and (96.75\% - 97.59\%) for accuracy. This range also includes the TransformerApt and TransformerCRF models, which were trained from scratch.

\begin{table}[!t]
\begin{tabular}{lcccc}
\toprule
\multicolumn{1}{c}{Model} & P     & R$^\dagger$     & F1\\ 
\midrule \midrule
BERT-Apt                    & 0.974 & 0.974 & 0.974\\
BERT-Crf                    & 0.974 & 0.974 & 0.974\\
\midrule
BioBERT-Apt                 & 0.745 & 0.863 & 0.800\\
BioBERT-Crf                 & \textit{0.974} & 0.975 & \textit{0.974}\\
\midrule
ClinicalBERT-Apt            & 0.974 & \textit{0.975} & 0.974\\
ClinicalBERT-Crf            & \textbf{0.976} & \textbf{0.976} & \textbf{0.975}\\
\midrule
TransformerApt              & 0.965 & 0.968 & 0.966\\
TransformerCRF              & 0.966 & 0.968 & 0.967\\
\bottomrule
\end{tabular}
\caption{BIO-strict weighted-evaluation results. The highest scores are in bold and the second highest in italic. $^\dagger$Equivalent to Accuracy and micro-F1.}
\label{tab:weighted_n_micro_scores}
\end{table}

Upon closer examination of the overall evaluation scores, we observe the following interesting findings:
\textbf{1)} Micro and Weighted Evaluations may not provide a comprehensive perspective. When dealing with scenarios where special labels have significantly low occurrences, the primary challenge for the model is to accurately label these entities. However, if the model merely assigns all tokens to the predominant class, such as "\textit{O}", it can still achieve remarkably high micro- and weighted-average evaluation scores. For instance, BioBERT-Apt exhibits macro precision/recall/F1 scores of 0.0454/0.0526/0.0488, but a micro-score of 0.8630 and weighted precision/recall/F1 of 0.7448/0.8630/0.7995; giving misleadingly high accuracies.
\textbf{2)} The introduction of CRF layers has an obvious impact. While in most instances, adding such a layer only leads to marginal improvements, BioBERT models experience a substantial shift in performance when implementing a CRF layer. The reasons for this significant difference remain subject to further investigation.

\subsubsection{BIO Classification Reports and Confusion Matrix of PLM-CRFs}
\begin{figure*}[!tbh]
\begin{center}
\centering
\includegraphics*[width=\textwidth]{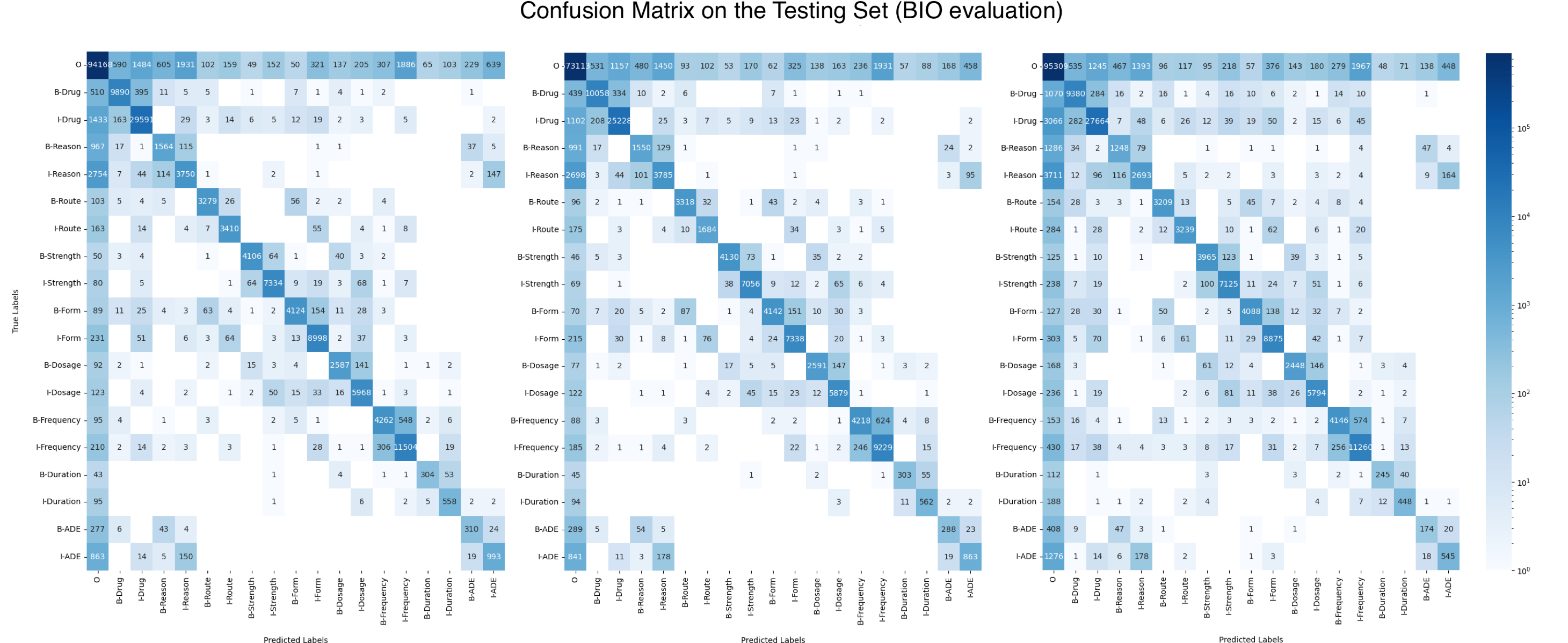}
\caption{BIO-strict Confusion Matrix of BERT-CRF (left), ClinicalBERT-CRF (middle) and TrasformerCRF (right)}
\label{fig:CRF-BIO-CONFM}
\end{center}
\end{figure*}

For a more in-depth analysis of the model performances on different labels, we provide a detailed breakdown of the BIO-strict evaluation scores in Figure \ref{fig:CRF-BIO} (classification report) and \ref{fig:CRF-BIO-CONFM} (confusion matrix). Since CRF models consistently outperform Adaptation models, we present the results of the following three representative models:
ClinicalBERT-CRF, as the best-performing model; BERT-CRF, representing our base LLM; and TransformerCRF, being our model trained from scratch.

When analysing the strict BIO evaluation scores for BERT-CRF and ClinicalBERT-CRF models (as depicted in Figure \ref{fig:CRF-BIO}), the following observations can be made:
Both models exhibit low performance at labelling ADE and Reasons, with F1 scores falling below 0.70.
While ClinicalBERT-CRF outperforms BERT-CRF in terms of F1 scores for B-Reason (0.63 vs. 0.62) and I-Reason (0.61 vs. 0.58), the two models achieve a tie for B-ADE (both at 0.49) with ClinicalBERT-CRF lagging slightly behind in I-ADE.
TransformerCRF exhibits a similar behaviour, while also registering an F1 score below 0.70 at B-Duration labelling.

Looking at the confusion matrix in Figure \ref{fig:CRF-BIO-CONFM}, we observe the following behaviors:
\begin{itemize}
    \item Most of the misclassified labels, predicted as out-of-context \textit{O} tokens, are I-Drug, I-Reason, and I-Frequency. Misclassification numbers are (1157, 1450, 1931) by Clinical-BERT-CRF, (1484, 1931, 1886) by BERT-CRF, and (1245, 1393, 1967) by TransformerCRF.
    \item Among the same special labels, the most frequent misclassifications happen between (I-Drug $\rightarrow$ B-Drug, B-Drug $\rightarrow$ I-Drug). Those are (208, 334) by Clinical-BERT-CRF, (163, 395) by BERT-CRF, and (282, 284) by TransformerCRF.
    \item Between different attributes, the most common cross-label misclassifications are (I-ADE $\rightarrow$ I-Reason, I-Reason $\rightarrow$ I-ADE). These are (178, 95) by Clinical-BERT-CRF, (150, 147) by BERT-CRF, and (178, 164) by TransformerCRF. The following pairs are (B-Form $\rightarrow$ B-Route, I-Form $\rightarrow$ I-Route, I-Strength $\rightarrow$ I-Dosage, B-ADE $\rightarrow$ B-Reason, and I-Dosage $\rightarrow$ I-Strength) for Clinical-BERT-CRF.
    
    In fact, these label confusions are coherent. For instance, \textit{ADE} and \textit{Reason}, or \textit{Dosage} and \textit{Strength} are known to be closely \textit{related} attributes, and often present some issues in NER tasks.
\end{itemize}

\subsection{Token-Level Evaluations}
\begin{figure*}[!tbh]
\begin{center}
\centering
\includegraphics*[width=\textwidth]{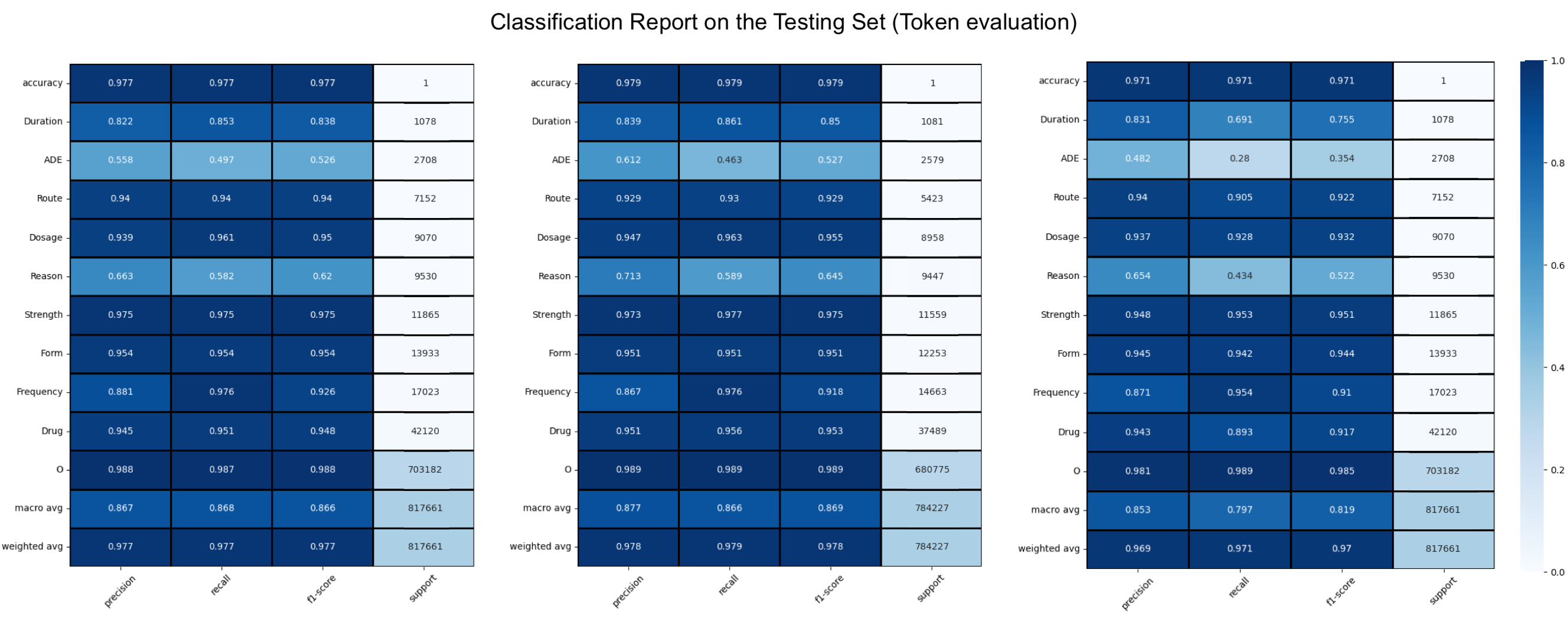}
\caption{Token-level Classification 
of BERT-CRF (left), ClinicalBERT-CRF (middle) and TrasformerCRF (right)}
\label{fig:CRF-TOKEN}
\end{center}
\end{figure*}

As mentioned in the evaluation setup, we also decide to report performance scores on non-detailed granularity (or token-level) - i.e. without distinguishing between the beginning (\textit{B}) and inner-part (\textit{I}) of an event. 
The resulting evaluation scores are reported in Figure \ref{fig:CRF-TOKEN}.
In comparison to the BIO-span level evaluation, we observe, without surprise, notably higher scores. Specifically, macro-F1 scores increased from (83.7 $\rightarrow$ 86.9), (83.72 $\rightarrow$ 86.6), and (78.74 $\rightarrow$ 81.9) for Clinical-BERT-CRF, BERT-CRF, and TransformerCRF respectively.

\subsection{Efficient Training Results at Token-level}
To investigate the results of utilising down-sampling to achieve better label distribution in the training and validation datasets, we present the evaluation outcomes of the three same models.

\begin{figure*}[!tbh]
\begin{center}
\centering
\includegraphics*[width=\textwidth]{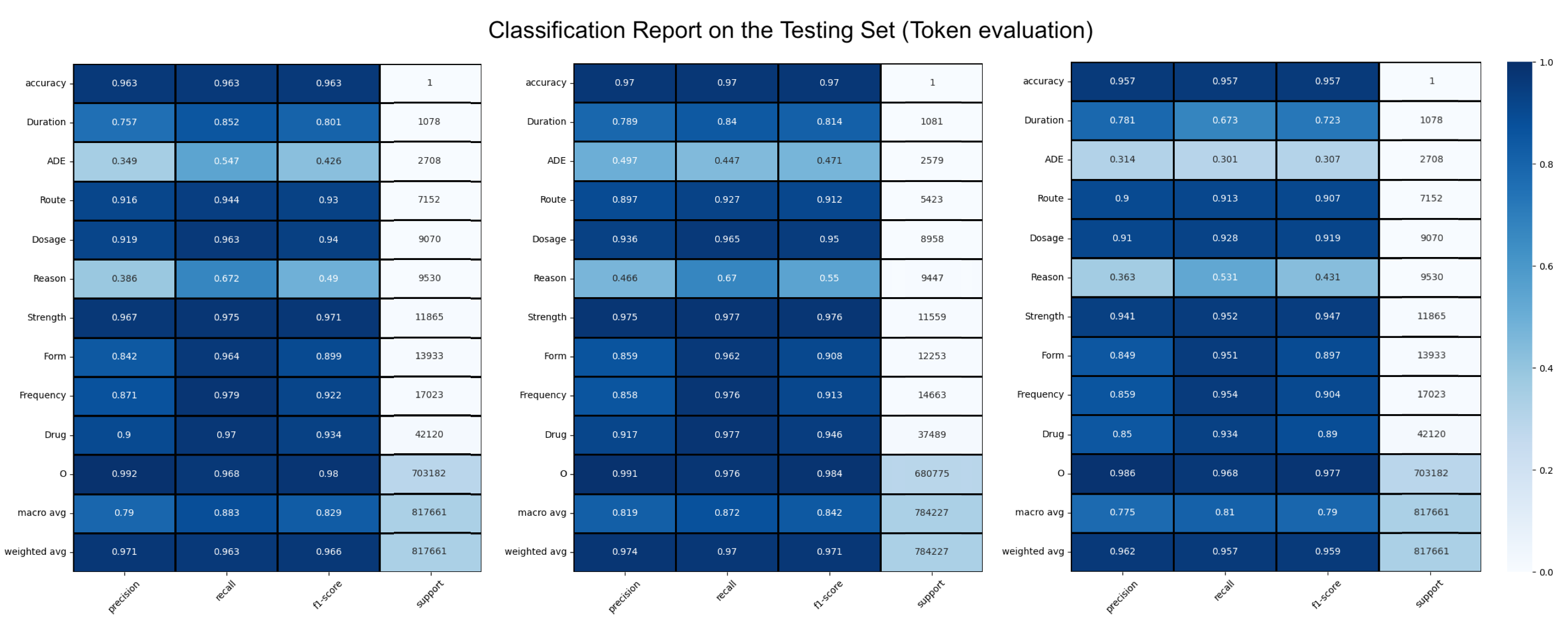}
\caption{Token-level Classification of BERT-CRF (left), ClinicalBERT-CRF (middle) and TrasformerCRF (right); using \textbf{Efficient Training}.}
\label{fig:CRF-TOKEN-SAMBD}
\end{center}
\end{figure*}

Figure \ref{fig:CRF-TOKEN-SAMBD} showcases the token-level evaluations when implementing efficient training on BERT-CRF, ClinicalBERT-CRF, and TransformerCRF.
Compared to the outcomes from training on the full dataset, we observe that efficient training yields very similar macro-average evaluation scores (0.829 to 0.866), (0.842 to 0.869), and (0.790 to 0.819). However, it accomplishes this with a reduced amount of data and computational cost. The weighted average F1 scores are even more closely aligned, ranging from 0.966 to 0.977, 0.971 to 0.978, and 0.959 to 0.97 respectively, for the fine-tuned PLMs BERT-CRF and Clinical-BERT-CRF, and trained-from-scratch TransformerCRF.

We deduce that, in experiments involving significantly larger datasets, efficient training could demonstrate even greater potential for saving on model training and parameter optimisation costs.

\subsection{Ablation Evaluations at Token-level}

To analyse the system performances in more detail, we carry out some ablation analysis and report three ablation evaluations at token-level, in Table \ref{tab:ablation-eval-o-ade-reason}.
\begin{itemize}
    \item Firstly, since the dataset is over-populated with ``O'' labels, which may bias the results of our models, we report new evaluation scores after the removal of ``O'' labels.
    \item Secondly, because the ``ADE'' and ``Reason'' labels are two outliers yielding much lower performances than other classes across all models, we decide to also conduct evaluation without these.
    \item Additionally, we evaluate the combination of the points above - that is, on the seven special labels excluding ``O'', ``ADE'' and ``Reason''. 
    \item Thirdly, because ``ADE'' and ``Reason'' are semantically related, we evaluate whether the models could make accurate predictions for tokens belonging to either of these classes. To do so, we merge and treat the "ADE" label as a "Reason" label, and leave all remaining labels unchanged (including ``O'' labels).
\end{itemize}

\begin{table*}[!h]
\begin{center}
\centering
\begin{tabular}{c|ccc|ccc|ccc}
\toprule
\multicolumn{1}{c}{Metrics} & \multicolumn{3}{c}{BERT-CRF }                  & \multicolumn{3}{c}{ClinicalBERT-CRF }   & \multicolumn{3}{c}{TransformerCRF (scratch) } \\
\midrule
&P&R&F&P&R&F&P&R&F \\ \hline 
\multicolumn{10}{c}{Removing label `O': only using 9 key labels}  \\\hline 
   micro avg   & 0.908 & \textbf{0.915}& 0. 912& \textbf{0.913}& 0.913 &\textbf{0.913} & 0.906& 0.863& 0.884 \\
   macro avg    & 0.853 & \textbf{0.854}&0.853 & \textbf{0.865} &0.852 & \textbf{0.856} & 0.839& 0.776 &0.801 \\
weighted avg     & 0.905& \textbf{0.915} &\textbf{0.91} & \textbf{0.909} &0.913 &\textbf{0.91} & 0.896& 0.863& 0.877 \\
\hline 
\multicolumn{10}{c}{Removing labels  `Reason' and `ADE': keeping `O' and other 7 key labels} \\\hline 
   micro avg  & 0.982& 0.983& 0.982 & 0.982& 0.985& 0.984 & 0.975& 0.98& 0.977 \\
   macro avg   & 0.931 &0.95 &0.94 & 0.931& 0.95& 0.94 & 0.925 & 0.907 &0.915\\
weighted avg    & 0.982 & 0.983& 0.983 & 0.983& 0.985& 0.984 & 0.975& 0.98& 0.977\\
\hline 
\multicolumn{10}{c}{Removing labels  `O', `Reason' and `ADE': only keeping other 7 key labels} \\\hline 
   micro avg  & 0.936& 0.957& 0.947& 0.936& 0.959& 0.947& 0.929& 0. 919& 0.924 \\
   macro avg   & 0.922 &0.944 &0.933 & 0.922& 0.945& 0.933 & 0.916 & 0.895&0.904\\
weighted avg    & 0.937 & 0.957& 0.947 & 0.937& 0.959& 0.948 & 0.93& 0.919& 0.924\\
\hline 
\multicolumn{10}{c}{ Merging two labels `ADE' as `Reason': keeping all others} \\\hline 
   macro avg   & 0.902 & 0.91& 0.906 & 0.908 &0.911& 0.908 &0.897& 0.855& 0.872 \\
weighted avg    & 0.977& 0.978& 0.977& 0.979& 0.979& 0.979 & 0.97 &0.972 &0.971\\
\bottomrule
\end{tabular}
\caption{Results of Ablation Evaluations at Token-level on the Original Full Dataset}
\label{tab:ablation-eval-o-ade-reason}
\end{center}
\end{table*}

From the results obtained in Table \ref{tab:ablation-eval-o-ade-reason}, we can observe that 
\textbf{1)} by removing ``O'' labels, the macro averaged F1 and weighted F1 scores increase to 0.853/0.856/0.801 and 0.910/0.910/0.887 for our three models. 
\textbf{2)} By removing the ``Reason'' and ``ADE'' outliers, we observe much higher macro- and weighted- F1 scores of 0.940/0.940/0.915 and 0.983/0.984/0.977 respectively.
\textbf{3)} By merging the labels ``ADE'' and ``Reason'', the scores also tend to increase.
\textbf{4)} Overall, the TransformerCRF model trained from scratch is about five points (0.05) below the other two models on the nine special labels, but only three points (0.03) below in terms of macro-F1 when merging ``ADE'' and ``Reason''.

\section{Revisiting n2c2-2018 Official Submissions }
\label{sec_discussion}

Looking back at the official results from the n2c2-2018 challenge, F1 scores achieved by the top-performing systems were all owing to the external knowledge-based features the teams used. That includes pre-trained word embeddings upon the entire MIMIC III data and part-of-speech (POS) tagging \cite{henry20202018_n2c2_task2}; involving high computational costs and resulting in models which may be hard to reproduce.
In addition, most of the submitted models implemented BiLSTM-CRF architectures. Instead, our study examined the latest BERT-like deep learning structures with CRF layers, using both pre-trained LLMs and TransformerCRF architecture. In fact, TransformerCRF is the only \textit{constrained models} that is restricted to only using the official training data of 303 annotated letters for training.
Nevertheless, TransformerCRF with and without efficient training (both) produced higher weighted average F1 scores at token-level (Lenient F1) with 0.970 and 0.959 (Fig. \ref{fig:CRF-TOKEN} and \ref{fig:CRF-TOKEN-SAMBD}) than the highest achieved F1 (0.942) from the official shared task submissions \cite{henry20202018_n2c2_task2}.
We call for attention from researchers on constrained clinical text mining and how to improve model performances when the available resources are low or restricted.



\section{Conclusion and Future Work}
\label{sec_conclude}

In this study, we delved into the following question: "Given a domain-specific task, how limited should the available data be for mixed-domain pre-training to be considered beneficial?". Leveraging a limited dataset of human-annotated clinical Electronic Health Records (EHR) comprising 303 letters and roughly 45,000 sentences from the n2c2 shared task, we examined the performance of Pre-trained Large Language Models in a clinical context. Specifically, we explored the use of BERT, BioBERT, and ClinicalBERT as representative examples of LLMs from general, biomedical, and clinical domains. We compared their fine-tuning with Adaptation and Conditional Random Field (CRF) layers against Transformer models trained from scratch (Transformer-Apt/CRF).

On the original training dataset, using macro-averaged F1 as our main metric, Clinical-BERT-CRF and BERT-CRF exhibited comparable top scores of 0.8371 and 0.8372 respectively, while BioBERT-CRF achieved 0.8357 and  TransformerCRF maintained a competitive score of 0.7839. Notably, TransformerApt and TransformerCRF featured significantly fewer parameters - i.e. 39.80\% less than BERT and BioBERT, and 49.51\% less than ClinicalBERT.
Furthermore, by implementing an efficient training method, we successfully reduced training costs by down-sampling against plain samples. This approach aimed at balancing label distributions across the training set while only reducing the overabundance of empty-labelled sentences, without significantly impacting performances.

In our ablation evaluations, although we observed that TransformerCRF performed marginally lower than PLMs in terms of Macro-F1 when considering all nine key labels, the performance gaps narrowed to a mere 0.03 when excluding "ADE" and "Reason". This result is particularly promising given the considerably lower computational cost of the TransformerCRF model trained from scratch.

Finally, the insights gained from this research prompt the following question: "Is fine-tuning LLMs a broadly optimal approach in Clinical NLP considering the limited and restricted nature of its data, or should training from scratch be considered the go-to approach in some specific scenarios?".

In future work, we plan to explore data augmentation techniques such as synthetic data generation \cite{LT3}, specifically on low-frequency labels to achieve further improvements. Additionally, we will investigate graph-based semi-supervised learning techniques, which involve graph construction on both labelled and unlabelled data through label propagation, as demonstrated by \cite{Han2015ChineseNE}.

\section*{Author Contributions}
S carried out the experimental work: pre-trained BERT, BioBERT and ClinicalBERT; trained the Transformer models; implemented Adaptation (Apt) and CRF extensions for each model; applied down-sampling (Efficient Training); and designed+ran experiments.
L designed the project.
S and L co-wrote the paper.
Y contributed to the earlier version of TransformerCRF with code and experimentation (\url{https://arxiv.org/abs/2210.12770v2}). 
G co-supervised the work.

\section*{Acknowledgements}
We thank Viktor Schlegel, Hao Li, Valerio Antonini, Ghada Alfattni for useful discussion on this project.
We thank Google's Colab service at our model developing stage and the CSF3 server support from The University of Manchester. 
We acknowledge the usage of the following packages: SKlearn, BERT, BioBERT, and ClinicalBERT.
The following projects have partially supported this research: UKRI/EPSRC grant
EP/V047949/1 “Integrating hospital outpatient letters into the healthcare data space”.

\bibliography{anthology,custom}
\bibliographystyle{acl_natbib}
\newpage
\appendix
\section*{Appendix}
\label{app-all}
We list the BIO-strict span-level evaluation reports and confusion matrices of BioBERT-CRF, as well as BERT-Apt, ClinicalBERT-Apt, and TransformerApt.

\clearpage

\begin{figure}[t!]
\begin{center}
\centering
\includegraphics*[width=\columnwidth]{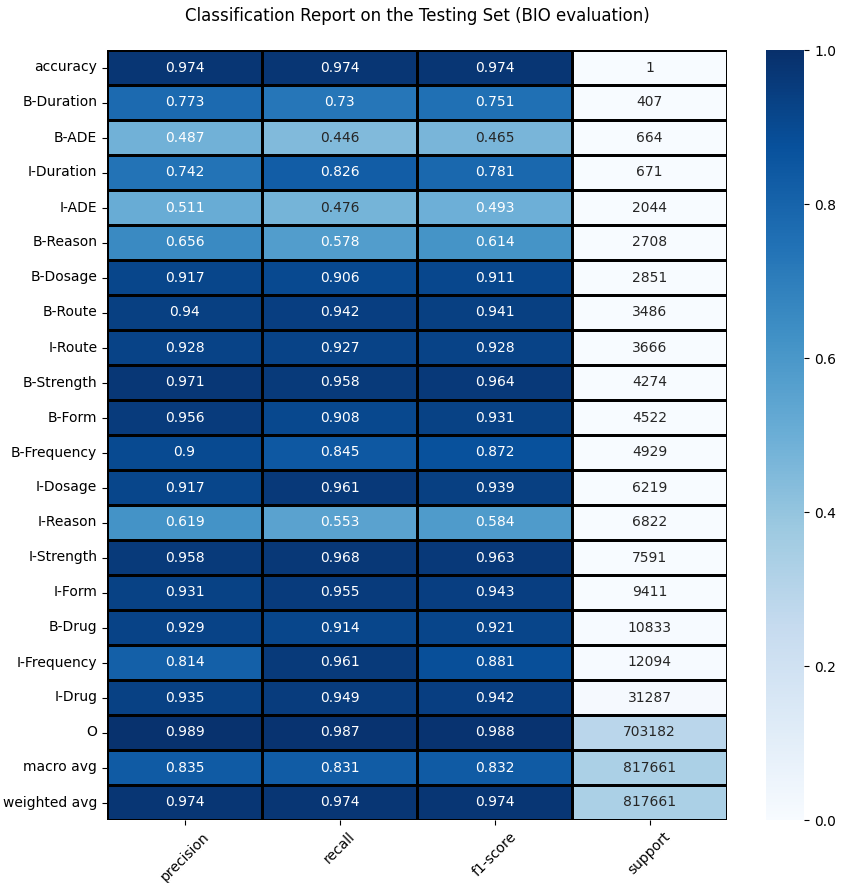}
\caption{BIO-strict Classification of BERT-Apt}
\label{fig:BERTapt_bio_classification_report}
\end{center}
\end{figure}

\begin{figure}[b!]
\begin{center}
\centering
\includegraphics*[width=\columnwidth]{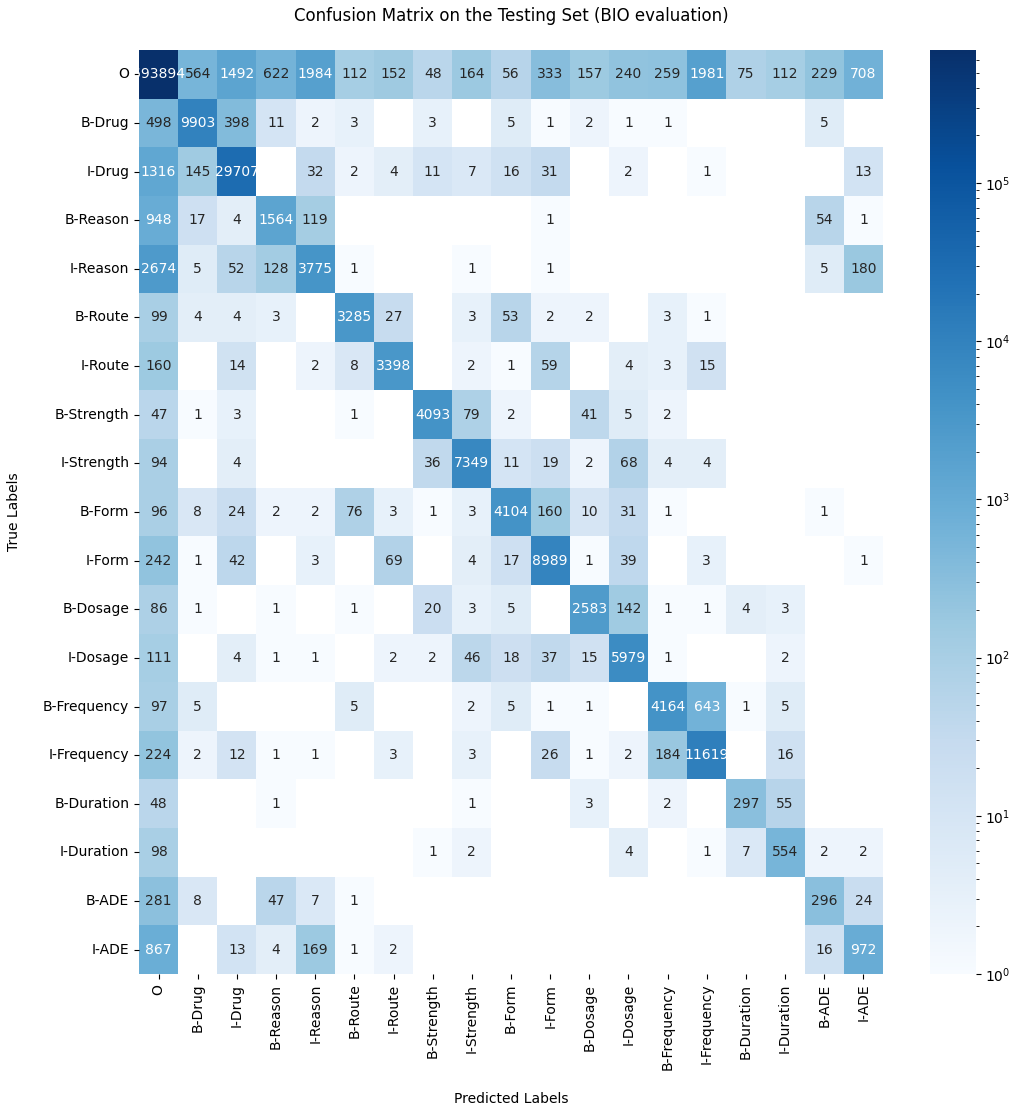}
\caption{BIO-strict Confusion Matrix of BERT-Apt}
\label{fig:BERTapt_bio_confusion_matrix}
\end{center}
\end{figure}

\begin{figure}[t!]
\begin{center}
\centering
\includegraphics*[width=\columnwidth]{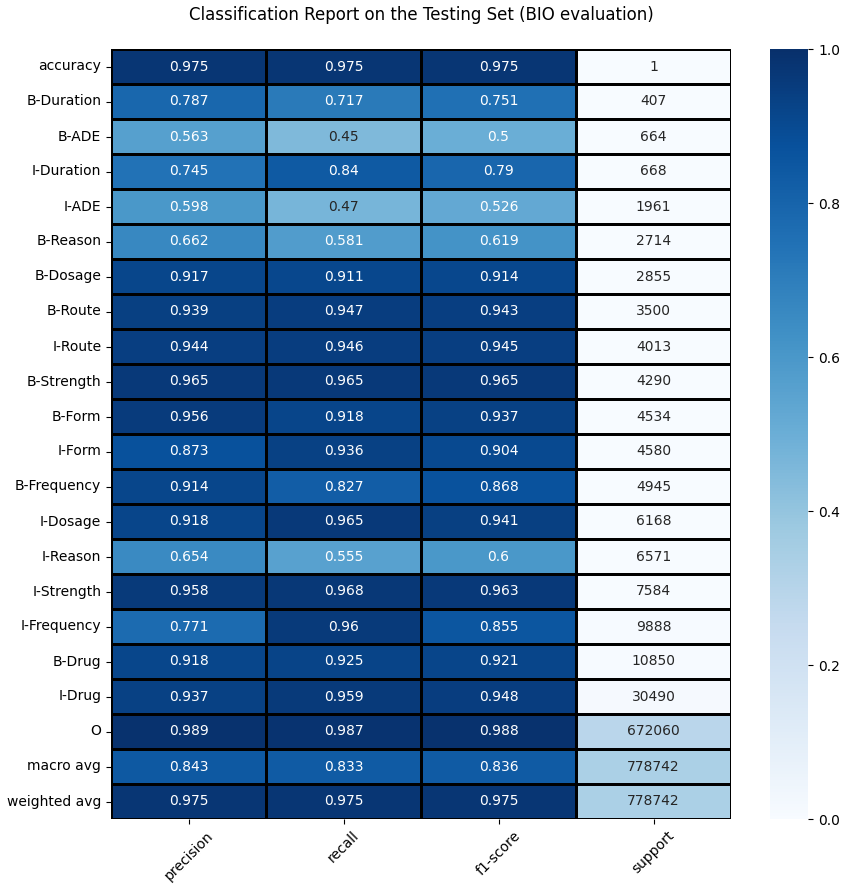}
\caption{BIO-strict Classification of BioBERT-CRF}
\label{fig:BioBERTcrf_bio_classification_report}
\end{center}
\end{figure}

\begin{figure}[b!]
\begin{center}
\centering
\includegraphics*[width=\columnwidth]{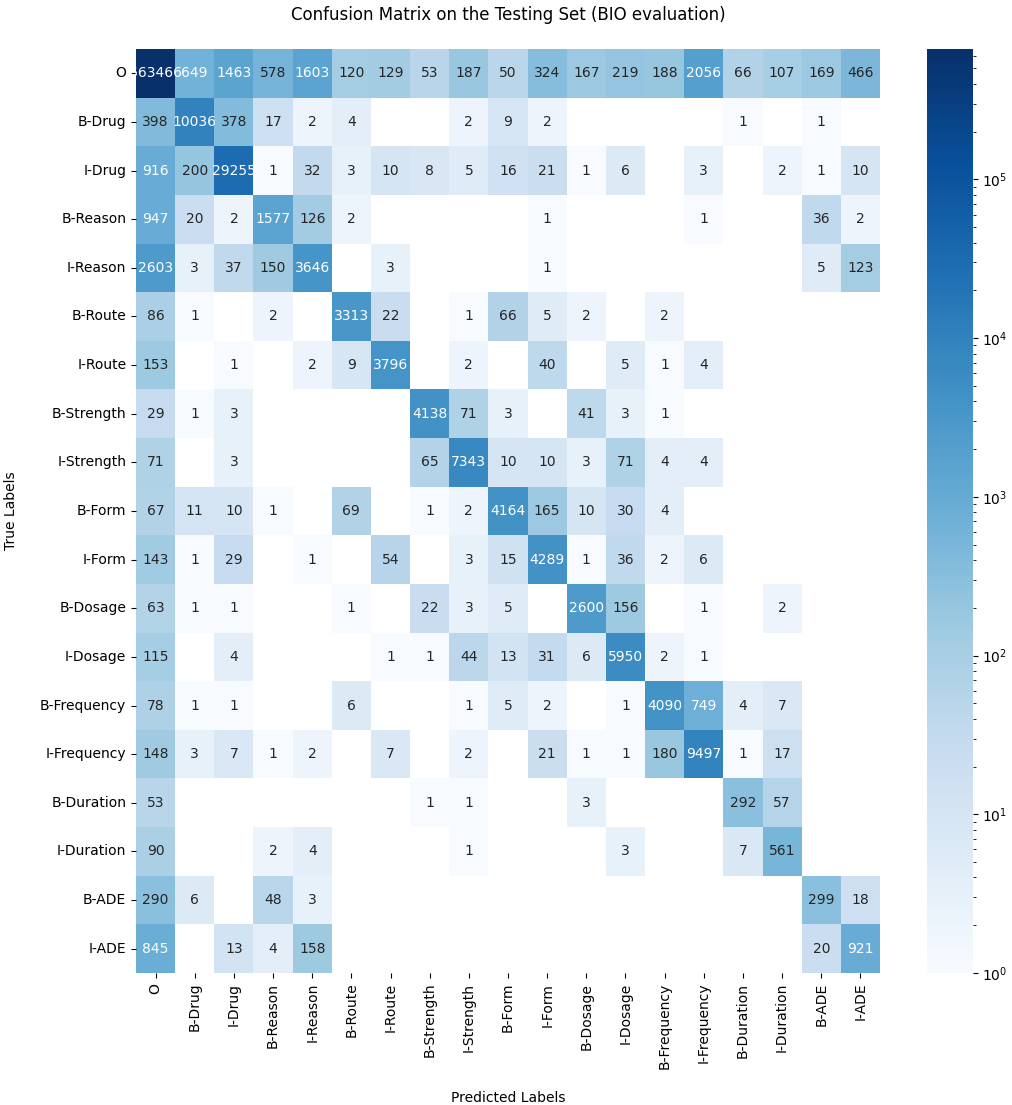}
\caption{BIO-strict Confusion Matrix of BioBERT-CRF}
\label{fig:BioBERTcrf_bio_confusion_matrix}
\end{center}
\end{figure}


\begin{figure}[t!]
\begin{center}
\centering
\includegraphics*[width=\columnwidth]{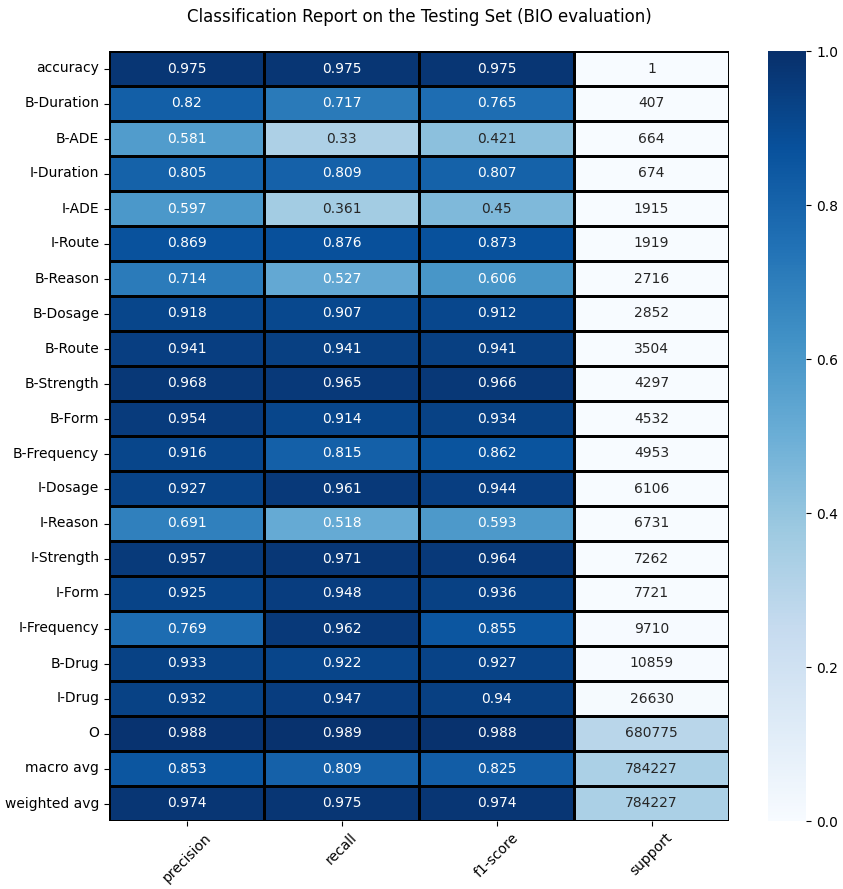}
\caption{BIO-strict Classification of ClinicalBERT-Apt}
\label{fig:ClinicalBERTapt_bio_classification_report}
\end{center}
\end{figure}

\begin{figure}[b!]
\begin{center}
\centering
\includegraphics*[width=\columnwidth]{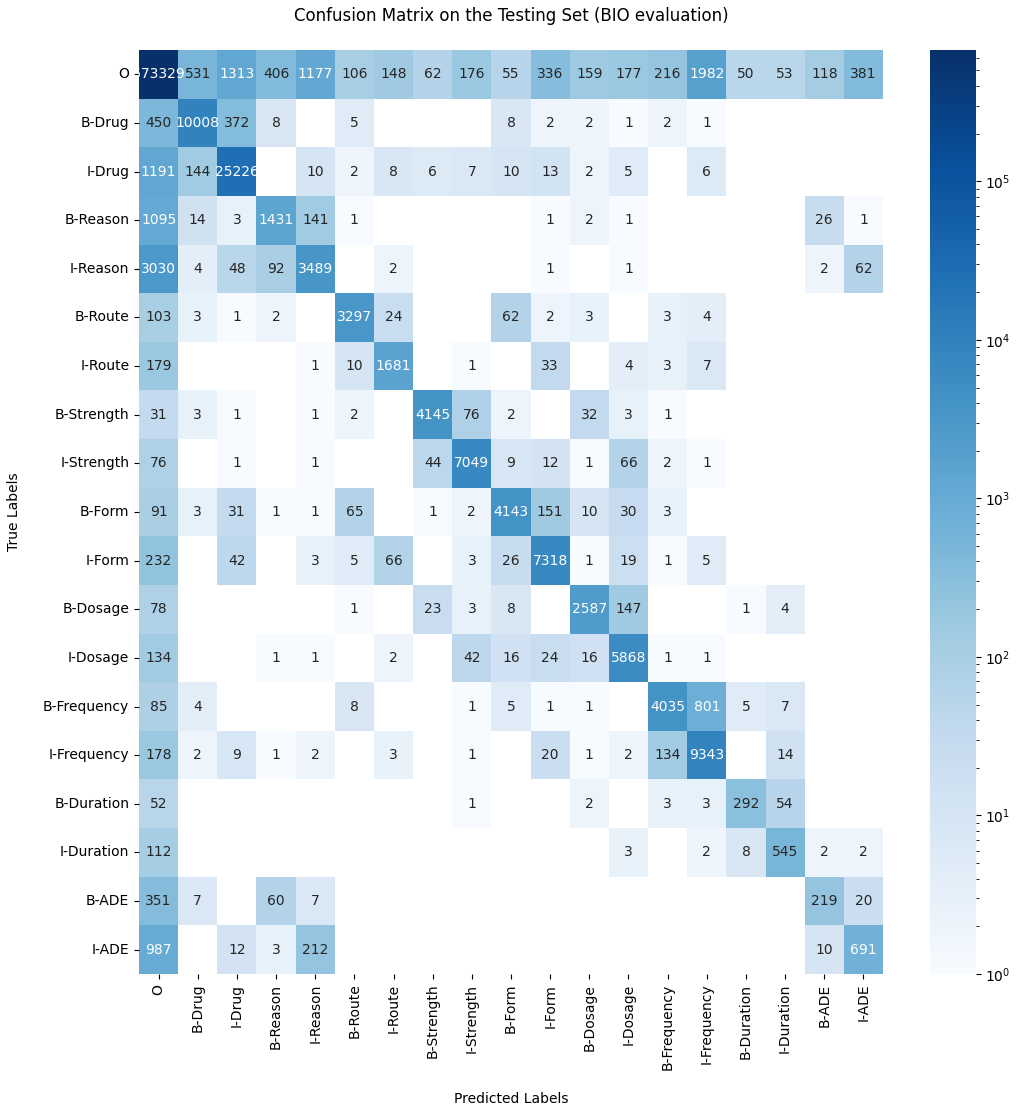}
\caption{BIO-strict Confusion Matrix of ClinicalBERT-Apt}
\label{fig:ClinicalBERTapt_bio_confusion_matrix}
\end{center}
\end{figure}

\begin{figure}[t!]
\begin{center}
\centering
\includegraphics*[width=\columnwidth]{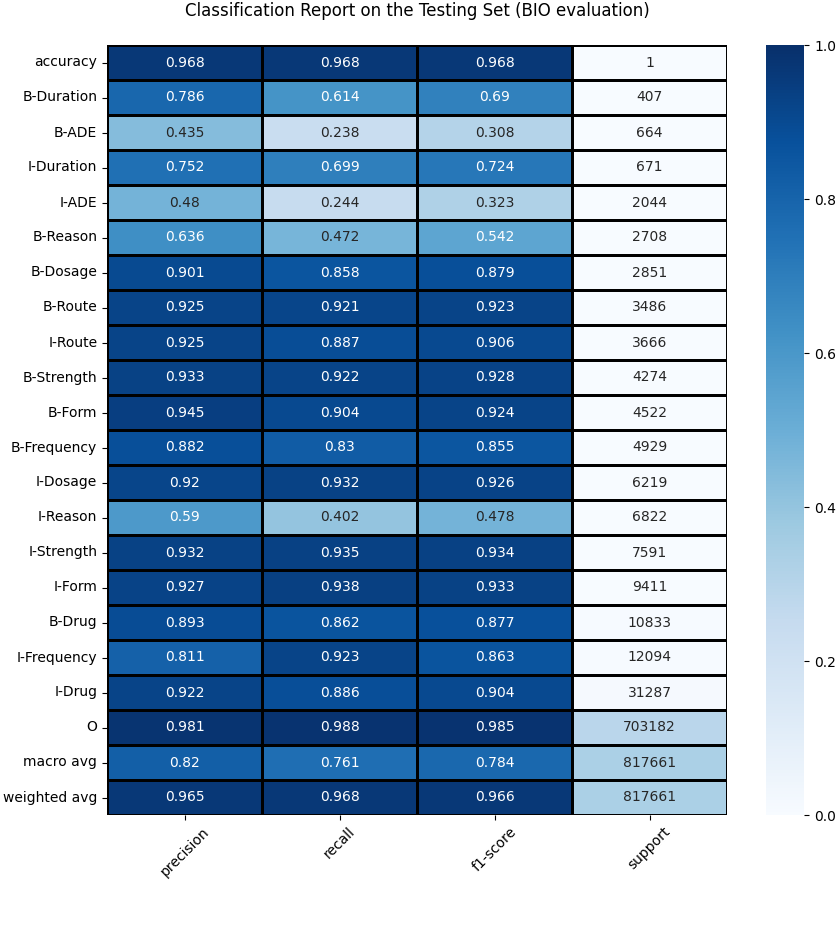}
\caption{BIO-strict Classification of TransformerApt}
\label{fig:TransformerApt_bio_classification_report}
\end{center}
\end{figure}

\begin{figure}[b!]
\begin{center}
\centering
\includegraphics*[width=\columnwidth]{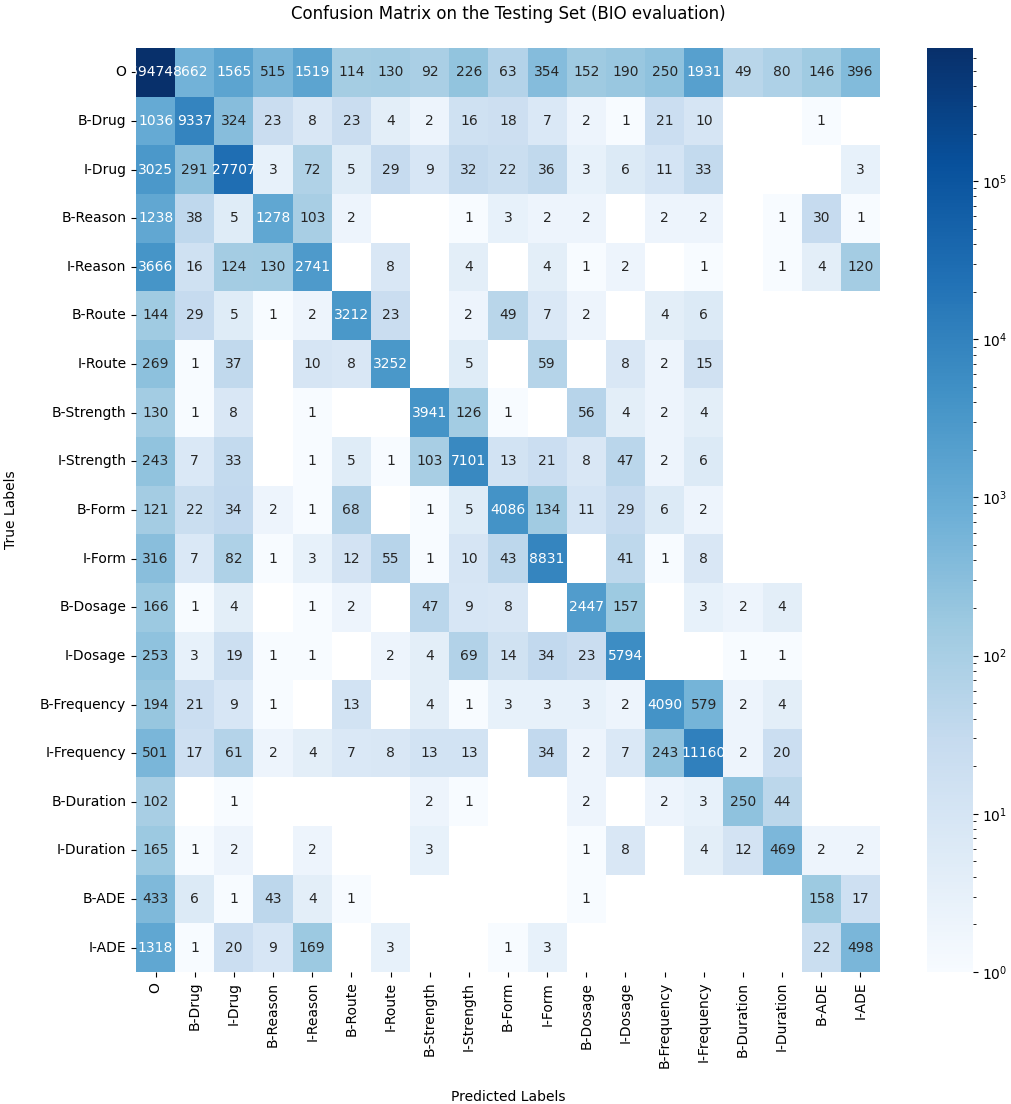}
\caption{BIO-strict Confusion Matrix of TransformerApt}
\label{fig:TransformerApt_bio_confusion_matrix}
\end{center}
\end{figure}

\end{document}